\documentclass[letterpaper]{article}
\PassOptionsToPackage{table}{xcolor}
\usepackage[draft]{aaai2026}
\usepackage{times}
\usepackage{helvet}
\usepackage{courier}
\usepackage[hyphens]{url}
\usepackage{graphicx}
\usepackage{natbib}
\usepackage{caption}
\usepackage{amsmath,amsfonts,amssymb}
\usepackage{array}
\usepackage{textcomp}
\usepackage{verbatim}
\usepackage{booktabs}
\usepackage{multirow}
\usepackage{xcolor}
\definecolor{best}{HTML}{e6dad1}

\title{Learning to Communicate Locally for Large-Scale Multi-Agent Pathfinding}
\author{
    Valeriy Vyaltsev, Alsu Sagirova, Anton Andreychuk, Oleg Bulichev, Yuri Kuratov, \\ Konstantin Yakovlev, Aleksandr Panov, Alexey Skrynnik
}
\affiliations{
    CogAI Lab, Moscow, Russia
}

\begin{document}

\maketitle

\begin{abstract}
Multi-agent pathfinding (MAPF) is a widely used abstraction for multi-robot trajectory planning problems, where multiple homogeneous agents move simultaneously within a shared environment. Although solving MAPF optimally is NP-hard, scalable and efficient solvers are critical for real-world applications such as logistics and search-and-rescue. To this end, the research community has proposed various decentralized suboptimal MAPF solvers that leverage machine learning. Such methods frame MAPF (from a single agent perspective) as a Dec-POMDP where at each time step an agent has to decide an action based on the local observation and typically solve the problem via reinforcement learning or imitation learning. We follow the same approach but additionally introduce a learnable communication module tailored to enhance cooperation between agents via efficient feature sharing. We present the Local Communication for Multi-agent Pathfinding (LC-MAPF), a generalizable pre-trained model that applies multi-round communication between neighboring agents to exchange information and improve their coordination. Our experiments show that the introduced method outperforms the existing learning-based MAPF solvers, including IL and RL-based approaches, across diverse metrics in a diverse range of (unseen) test scenarios. Remarkably, the introduced communication mechanism does not compromise LC-MAPF's scalability, a common bottleneck for communication-based MAPF solvers.
\end{abstract}

\section{Introduction}
Modern robotic systems often involve multiple mobile agents that must navigate and operate within shared environments, such as robots transporting goods in automated warehouses~\cite{li2021lifelong} or autonomous vehicles on public roads~\cite{li2023intersection}. A key abstraction for modeling and solving the challenge of coordinating such agents safely is multi-agent pathfinding (MAPF)~\cite{stern2019multi}.

In MAPF, time is divided into discrete steps, and agents move on a graph (typically a 4-connected grid). Agents act synchronously; at each step, each agent either moves to a neighboring vertex or waits in place. The goal is to compute a set of individual plans, one for each agent, that ensures no collisions occur during execution.

\begin{figure}[!t]
    \centering
    \includegraphics[width=\linewidth]{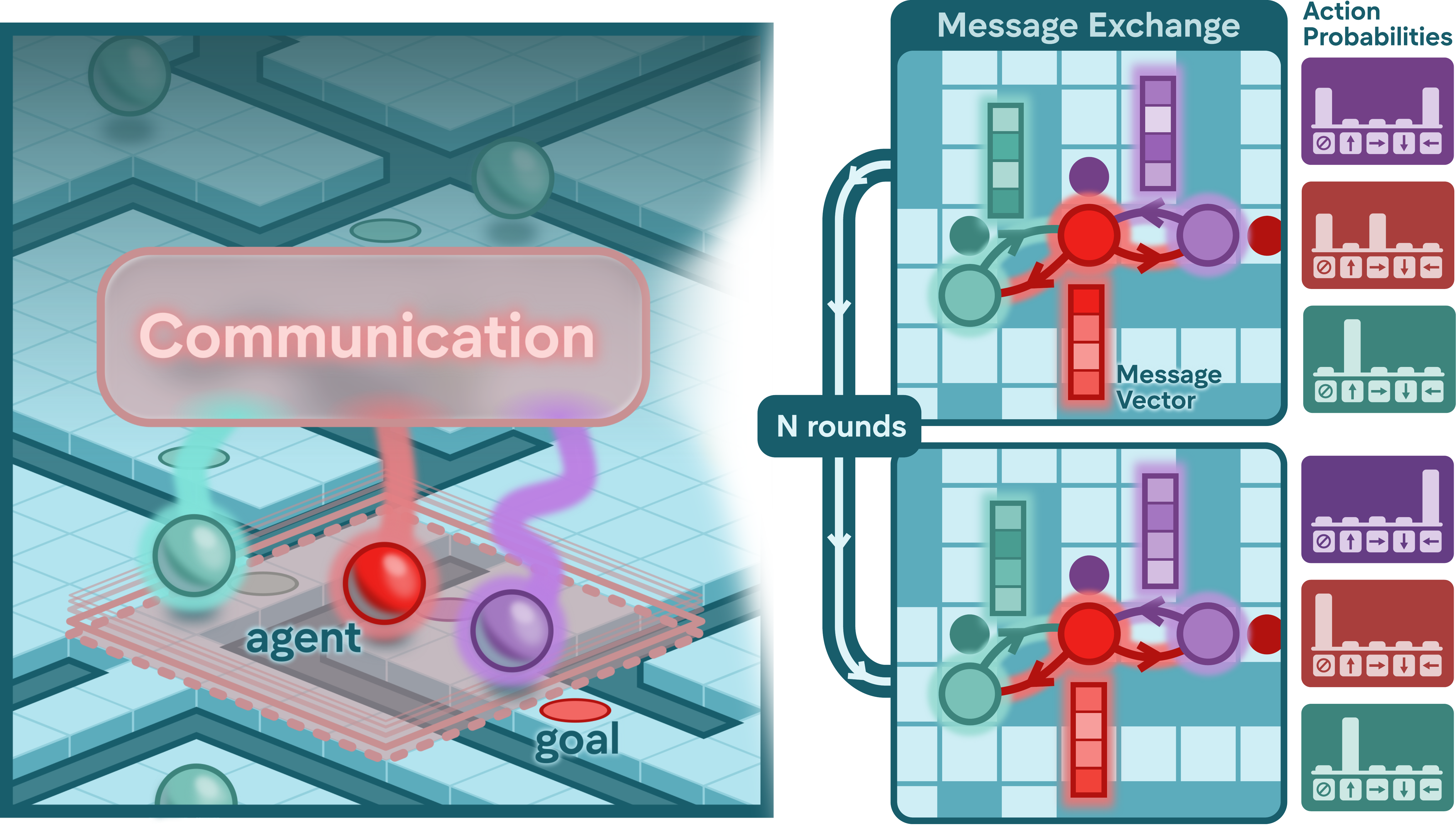}
    \caption{Iterative message exchange among agents enables progressive refinement of predicted action distributions over multiple communication rounds.}
    \label{fig:graphical_abstract}
\end{figure}

Many challenges of real-world robotics are not directly captured by the MAPF abstraction, including continuous space and time, asynchronous agent behavior, limited communication and observation, and various perception constraints. Despite these simplifications, MAPF successfully models the central difficulty in multi-robot navigation: coordinating agents to avoid collisions while aiming to optimize a specific cost function. As a result, MAPF has attracted substantial interest from both the robotics and AI research communities. Furthermore, a number of studies have demonstrated the successful application of MAPF-based methods to the continuous, noisy, and uncertain environments faced by real-world robotic systems~\cite{honig2016multi,ma2019lifelong}.

MAPF is most commonly approached in a centralized setting, where a single planner with full knowledge of the environment is responsible for generating plans for all agents. A wide range of both optimal and suboptimal centralized solvers have been proposed~\cite{standley2010finding,sharon2015conflict,Wagner2011,surynek2016efficient,okumura2022priority,okumura2023lacam,li2022mapf, veerapaneni2024improving, wang2025lns2}.

It is well established that optimal MAPF solvers scale poorly as the number of agents increases, as the problem is NP-hard~\cite{surynek2010optimization}. Suboptimal solvers, on the other hand, can scale to thousands of agents, but their solution quality may degrade significantly. Consequently, a central focus of MAPF research is striking a balance between computational efficiency and solution quality.

One promising strategy for addressing this challenge is to adopt a decentralized approach. Here, MAPF is modeled as a decentralized sequential decision-making problem, where each agent independently selects and executes actions at every time step based on local observations. The resulting decision-making policy may be fully learned or designed as a hybrid, combining learnable and fixed components~\cite{liu2020mapper,li2021message,wang2023scrimp,ma2021distributed,ma2021learning,tang2024ensembling,skrynnik2023learn,skrynnik2023switch,sagirova2025srmt, phan2025generative}. A recent survey provides a comprehensive overview of developments in this area~\cite{alkazzi2024comprehensive}.

A recent advancement in decentralized, learnable MAPF is MAPF-GPT~\cite{andreychuk2025mapf}, which relies entirely on supervised learning using a transformer-based neural network trained on an extensive dataset of approximately one billion observation-action pairs. Despite its simplicity, MAPF-GPT outperforms numerous other state-of-the-art learning-based MAPF methods.

However, a major limitation of MAPF-GPT is its lack of agent-to-agent communication. While it learns collaborative behavior from data, it does so without communication between agents, as the training data is generated by a centralized solver that does not provide communication signals. Consequently, MAPF-GPT cannot explicitly facilitate interaction or collaboration between agents during problem solving, which could be a key factor in improving performance.

Several existing decentralized MAPF methods, such as MAGAT~\cite{li2021message}, SCRIMP~\cite{wang2023scrimp}, DHC~\cite{ma2021distributed}, and DCC~\cite{ma2021learning}, use communication mechanisms. However, this communication is mostly limited to sharing local observations or internally known state information in one round~\cite{alkazzi2024comprehensive}. These mechanisms often fall short of enabling agents to engage in more meaningful coordination.

We argue that effective communication in decentralized MAPF should extend beyond single-message exchange and involve multiple rounds of agent interaction. Such iterative communication enables agents to negotiate, resolve conflicts, and build consistent joint plans that are crucial for robust multi-agent coordination in complex environments.
Motivated by this, we explore how to equip a large transformer-based imitation learning model with the ability to perform effective local communication.

Our main contributions are the following:

\begin{itemize}
\item We introduce a novel communication learning framework called \textbf{LC-MAPF}, which enables agents to communicate using only the expert demonstrations of selected actions, without requiring explicit communication supervision.

\item We present a transformer-based model with 3 million parameters that significantly improves performance and sets a new state-of-the-art among learnable decentralized MAPF solvers. We conduct extensive evaluations and compare it with existing learning-based approaches.
\item We extensively study how the number of communication rounds influences agent performance. We also show that, despite incorporating communication, the proposed mechanism maintains linear scalability as the number of agents grows.
\end{itemize}

\section{Related Work}

We discuss three categories of related work relevant to the proposed approach: foundation models for multi-agent systems, communication-based learning in MAPF, and multi-agent pathfinding.

\subsection{Foundation Models for Multi-Agent Systems}
Foundation models are typically trained on large-scale datasets, enabling generalization through zero-shot or few-shot learning~\cite{bommasani2021opportunities,yang2023foundation}.
For autonomous agents, demonstrations of task execution in the environment are used as a dataset, and generalization implies executing new tasks outside the training distribution without additional demonstrations or with only a small number of them~\cite{firoozi2023foundation}. Demonstration-based pretraining is not commonly used in multi-agent settings, but there are some examples, including games such as chess~\cite{silver2016mastering, ruoss2024amortized}, cooperative video games via self-play~\cite{berner2019dota}, and multi-agent pathfinding, as in SCRIMP~\cite{wang2023scrimp}.

A key strength of foundation models is their fine-tuning capability, which supports rapid adaptation to task-specific requirements. While widely adopted in robotics, particularly in multimodal tasks involving text-based instructions~\cite{firoozi2023foundation,team2024octo,kim2024openvla}, their use in multi-agent systems remains relatively limited. Notable examples include the Magnetic-One model for language and multimodal tasks in WebArena~\cite{fourney2024magentic} and MAPF-GPT for decentralized pathfinding~\cite{andreychuk2025mapf}.

\subsection{Multi-agent Pathfinding}

A variety of approaches have been proposed for solving MAPF. Rule-based solvers are designed for fast computation but lack guarantees on solution quality~\cite{okumura2023lacam,li2022mapf}. Reduction-based methods convert MAPF into classical problems such as minimum-cost flow or SAT, leveraging existing solvers to compute optimal solutions~\cite{surynek2016efficient}. Search-based solvers, such as CBS and its variants~\cite{sharon2015conflict,sharon2013increasing,Wagner2011}, apply graph search techniques and often offer optimality or bounded-suboptimality guarantees. Simpler methods like prioritized planning~\cite{ma2019searching} trade off optimality for efficiency and scalability.

\subsection{Communication-based MAPF Methods}

More recently, learning-based approaches have emerged. PRIMAL~\cite{sartoretti2019primal} was among the first to demonstrate decentralized MAPF solving via learning. In PRIMAL, the only communication between agents consists of their corresponding targets. One of the first learnable MAPF solvers with a dedicated learnable communication block was DHC~\cite{ma2021distributed}, which demonstrated significant improvement over PRIMAL. Subsequent methods such as DCC~\cite{ma2021learning} build on DHC, but enhance the communication mechanism by learning selective communication. Another approach, SCRIMP~\cite{wang2023scrimp}, combines imitation learning, reinforcement learning, and communication mechanisms, improving efficiency even further.

Another line of communication-based methods builds on graph attention networks. MAGAT~\cite{li2021message} replaces fixed communication structures with a graph attention mechanism, allowing agents to dynamically weight messages from neighbors for stronger coordination. Recently, multiple improved variants have appeared: HMAGAT~\cite{jain2026pairwise}, which utilizes hypergraphs, and MAGAT+~\cite{jain2025graph}, an enhanced version of MAGAT that uses three stacked graph attention layers compared to the single layer in the original.


\section{Background}

\subsection{Problem Statement}
A MAPF problem is a tuple $(G, s^1, ..., s^n, g^1, ..., g^n)$, where $G=(V,E)$ is a graph representing the environment, $s^i\in V$ is the start vertex of the $i$-th agent, and $g^i \in V$ is its goal vertex. A total of $n$ agents ($\mathcal{A} = \{u_1, \dots, u_n\}$) are present in the environment. The task is to find a set of plans $Pl = \{pl^i\}$ whose actions either move an agent to an adjacent vertex or keep it at its current vertex. Additionally, the plans should be conflict-free, i.e., no two agents occupy the same vertex or traverse the same edge at the same time. The solution cost is typically measured by either the \textit{Sum-of-Costs}, $SOC(Pl) = \sum_{i=1}^n cost(pl^i)$, or the \textit{makespan}, $msn(Pl) = \max_{i=1}^n cost(pl^i)$, where $cost(pl^i)$ is the timestep at which agent $i$ reaches its goal and remains there.

MAPF can also be formulated as a sequential decision-making problem, where the task is to construct a centralized policy $\pi_{\text{centralized}}$ that selects a joint, conflict-free action $\textbf{a} = a^1 \times \dots \times a^n$ at each timestep, with $a^i$ denoting agent $i$'s action. Such a policy can be hand-crafted or learned.

Finally, MAPF can also be treated as a decentralized decision-making problem where the goal is to learn a homogeneous individual policy $\pi$ shared by all agents, which selects an action for each agent based solely on local observations and, possibly, communication. The observations typically include information about nearby obstacles and agents, rather than the full global state.

\subsection{Imitation Learning for MAPF}
Imitation learning seeks to approximate an expert policy $\hat{\pi}$ by training a parameterized policy $\pi_\theta$. A dataset $\mathcal T$ of expert trajectories is first collected: $\hat{\mathcal{T}} = \{ \hat{\tau}_i \}_{i=1}^{K},$ where each $ \quad \hat{\tau}_i = \{ (s^1, \mathbf{a}^1), \dots, (s^L, \mathbf{a}^L) \}$ consists of state and joint action pairs. In MAPF, $\hat{\pi}$ is typically a centralized solver, for example, LaCAM*~\cite{okumura2024engineering}.

To enable decentralized learning, individual agent trajectories $\tau_u^{\hat{\pi}} = \{ (o^1_u, a^1_u), \dots, (o^L_u, a^L_u) \}$ are extracted, where $o^t_u$ is the local observation of agent $u$ at time $t$, and $a^t_u$ is the corresponding expert action. Observations may be represented as tensors or token sequences (e.g., in transformer-based models~\cite{ruoss2024amortized}). The resulting dataset $\mathcal{D} = \{ \tau_u^{\hat{\pi}} \}_{u=1}^{n}$ is then used to train the policy.

The learning objective minimizes the negative log-likelihood of expert actions:

\begin{equation}
\theta^\star = \arg\min_{\theta} \mathbb{E}_{(o_u, a_u^{\hat{\pi}}) \sim \mathcal{D}} \left[ -\log \pi_\theta (a_u^{\hat{\pi}} \mid o_u) \right].
\label{eq:cross-entropy}
\end{equation}

After training, actions are sampled as $a^u \sim \pi_\theta(o_u)$.

\begin{figure*}[!t]
    \centering
    \includegraphics[width=0.9\linewidth]{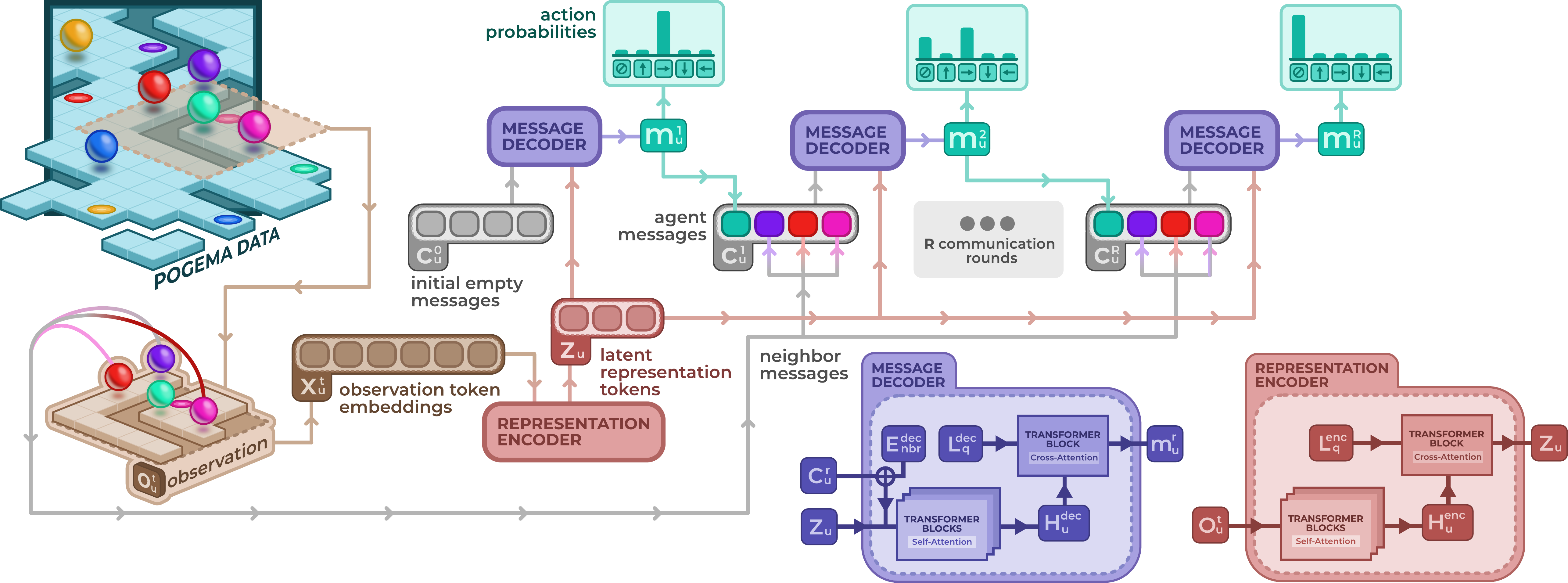}
    \caption{%
    Overview of the proposed LC-MAPF architecture.
Each agent $u \in \mathcal{U}$ encodes its local observation $o_{u}^{t}$ into a latent representation $z_{u}$ using a Transformer-based encoder. The latent state participates in an iterative message-passing procedure over $R$ communication rounds. At each round $r$, the agent receives the set of neighbor messages $C_{u}^{r}$ and fuses them with its latent state $z_{u}$ through a Transformer-based decoder, producing an updated message $m_{u}^{r}$ that will be sent to its neighbors in the next round. After $R$ communication rounds, the decoder outputs action logits $a_{u}$, which are converted to an action probability distribution $p_{u} = \mathrm{softmax}(a_{u})$. Both encoder and decoder consist of stacked Transformer blocks with self-attention and cross-attention: the encoder integrates the tokenized observation $o_{u}^{t}$, while the decoder integrates the agent's latent representation $z_{u}$ with contextualized neighbor messages $C_{u}^{r}$, enabling decentralized coordination through end-to-end learned communication.
}
    \label{fig:lc-mapf}
\end{figure*}

\section{Method}

The overall communication and action prediction workflow is illustrated in Fig.~\ref{fig:lc-mapf}.
At each time step $t \in [1,\dots, L]$ and for each agent $u \in [1,\dots, U]$, the model receives a structured observation
\begin{equation}
o_{u}^{t} = [\text{cost-to-go}_{u}^{t},\, i_{u}^{t},\, n_{u,1}^{t}, \dots, n_{u,k}^{t}],
\end{equation}
where $\text{cost-to-go}_{u}^{t}$ is an egocentric cost-to-go matrix, $i_{u}^{t}$ contains the agent's own features
(relative positions of current and goal locations, greedy action, and previous $k$ actions), and each
$n_{u,j}^{t}$ contains analogous information for one of the $k$ closest neighboring agents. This localized,
tokenized representation allows each agent to reason using only information that is relevant for preventing
collisions and coordinating movement.

The observation is converted into a sequence of embeddings:
\begin{equation}
    X_{0,u} = E_{\text{tok}}(o_{u}^{t}) + E_{\text{pos}} + E_{\text{nbr}},
\end{equation}
where $E_{\text{tok}}$ encodes the token identity, $E_{\text{pos}}$ provides a positional index inside the
sequence, and $E_{\text{nbr}}$ serves as a neighbor identifier so that the model can distinguish which
nearby agent contributed each token.

A Transformer encoder processes this embedded input and produces contextualized representations:
\begin{equation}
    H_{u}^\text{enc} = \mathrm{Encoder}(X_{0,u}).
\end{equation}

To avoid propagating the entire observation sequence throughout communication, we apply an information
bottleneck inspired by the Perceiver architecture \cite{jaegle2022perceiveriogeneralarchitecture}. A small
set of learnable latent queries $L_q^{enc}$ cross-attends to the encoded tokens and produces a compact latent state:
\begin{equation}
    z_{u} = \mathrm{LatentEncoder}(L_q^{enc}, H_{u}^\text{enc}),
    \qquad z_{u} \in \mathbb{R}^{T_{\text{latent}} \times d_{\text{latent}}}.
\end{equation}
This forms the agent's internal representation of the world and only this compressed state participates in
communication, making the communication cost independent of the observation size.
The agents then perform $R_{\text{comm}}$ rounds of local communication.
Each agent stores a message vector $m_{u}^{r} \in \mathbb{R}^{d_{\text{latent}}}$ for round $r$, initialized with
a learnable vector $m_{u}^{0}$ shared across all agents.

At round $r$, agent $u$ receives messages from neighbors:
\begin{equation}
    C_{u}^{r} = \{\, m_{v}^{r-1} + E_{\text{nbr}}^{dec}(v) \,\}_{v \in \mathcal{N}(u) \cup \{u\}},
\end{equation}
where $E_{\text{nbr}}^{dec}$ indicates which agent produced each message.
Messages are inserted into a decoding module together with the agent's latent state:
\begin{equation}
    h_{u}^{r} = \mathrm{Decoder}(L_q^{dec}, [z_{u}, C_{u}^{r}]).
\end{equation}
The decoder integrates information from neighbors and produces a new message:
\begin{equation}
    m_{u}^{r} = \mathrm{MsgHead}(h_{u}^{r}).
\end{equation}

After the final round, a prediction head produces the action logits:
\begin{equation}
    a_{u} = \mathrm{ActionHead}(h_{u}^{R_{\text{comm}}}), \qquad
    p_{u} = \mathrm{softmax}(a_{u}).
\end{equation}

Our Transformer blocks integrate several recent advancements aimed at improving stability and performance, including RMSNorm \cite{zhang2019rootmeansquarelayer} for normalization, SwiGLU \cite{shazeer2020gluvariantsimprovetransformer} feed-forward layers, a combined pre- and post-normalization and QK-normalization scheme \cite{zhuo2025hybridnormstableefficienttransformer}, and a differential attention mechanism \cite{ye2025differentialtransformer}.

LC-MAPF is trained from expert demonstrations in a fully end-to-end manner.
The training objective is the cross-entropy loss:
\begin{equation}
    \mathcal{L} =  \text{CE}\left(a_{u}, a_{u_*}\right)
\end{equation}
where $a_{u_*}$ is the one-hot expert action.
The total loss is averaged across all agents in the batch.

A key property of LC-MAPF is that messages are not supervised.
There is no auxiliary loss on $m_{u}^{r}$, nor are the messages forced to represent explicit semantic content.
Instead, messages influence future rounds of communication, and therefore their gradients flow through the
action loss of the agents that receive them. During backpropagation, the update to $m_{u}^{r}$ depends on how
it affects the action logits of neighboring agents in subsequent rounds. Consequently, the network learns what information should be communicated, allowing communication to emerge naturally from optimization of the shared objective.
In all our experiments, we use $R_{\text{comm}} = 4$ communication rounds.

\section{Experimental Setup}

\begin{figure*}[!t]
    \centering
    \includegraphics[width=0.95\textwidth]{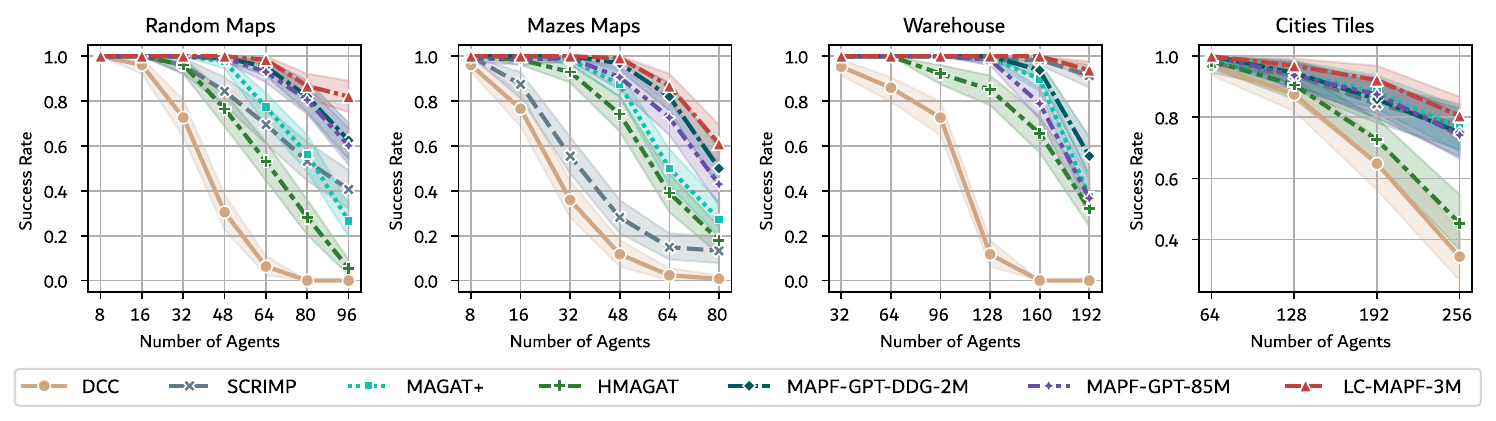}
    \caption{Success rates of the approaches on different map types depending on the number of agents in the instances (higher is better). The shaded area indicates the 95\% confidence interval.}
    \label{fig:csr_plot}
\end{figure*}

\begin{figure*}[!t]
    \centering
    \includegraphics[width=0.95\textwidth]{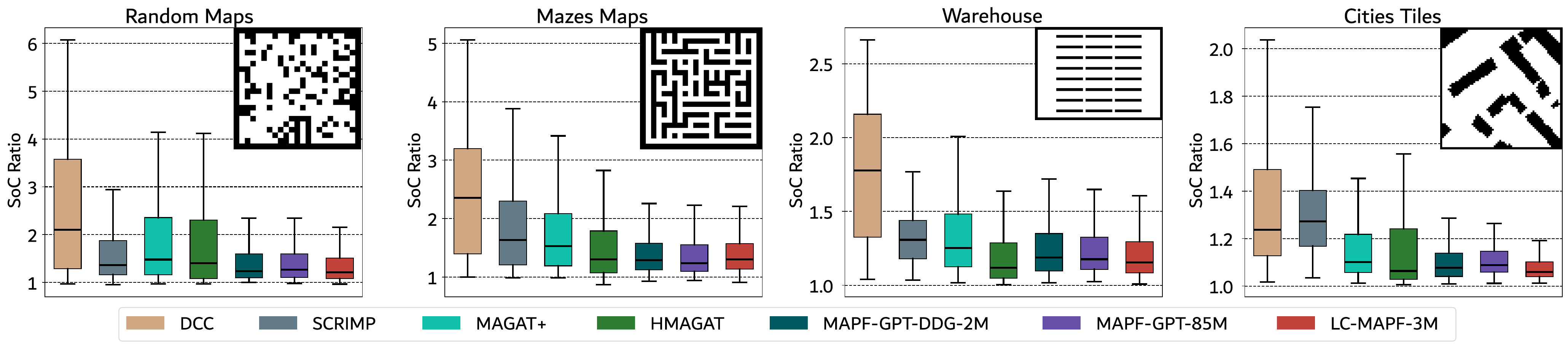}
    \caption{SoC ratio relative to solutions found by the LaCAM* approach (lower is better).}
    \label{fig:soc_plot}
\end{figure*}

The experiments were conducted on the \textbf{POGEMA} benchmark~\cite{skrynnik2025pogema},
which provides a diverse set of partially observable multi-agent pathfinding (MAPF) environments,
including \textit{Random}, \textit{Mazes}, \textit{Warehouse}, and \textit{Cities} map types.
Each agent receives a tokenized observation of up to \textbf{256 tokens}, corresponding to its $11\times11$ local field of view,
agent-specific attributes, and spatial context.
Each agent receives up to 13 messages within a 5-cell radius, including its own message,
and neighboring agents are ordered by their distance to the ego-agent,
ensuring consistent positional ordering in the communication graph.

The training dataset consisted of aggregated samples from three subsets of the POGEMA benchmark:
\textit{mazes}, \textit{random}, and \textit{house} maps.
The combined dataset contained approximately \textbf{23.5 million samples} with a 0.6:0.2:0.2 distribution across the three subsets.
In contrast to the dataset used for training the original MAPF-GPT, each sample in this dataset contains observations and ground-truth actions for all agents, rather than for a single selected agent. Thus, the total number of observation-action pairs is roughly \textbf{750 million}. In addition to the observations and ground-truth actions, the dataset contains adjacency information describing the local communication structure.

Training was performed for \textbf{800,000 iterations} using a single NVIDIA H100 GPU with 80GB memory.
Each iteration used a local batch of 32 samples with 16 gradient accumulation steps,
resulting in an effective batch size of 512 samples per optimization step.
The total training time amounted to roughly \textbf{900 GPU-hours}.
Mixed-precision training (bfloat16) was used to improve throughput and reduce memory footprint,
and all runs were executed using PyTorch~2.3.1 with CUDA~12.2.

The model contained approximately \textbf{3 million trainable parameters} and was trained from scratch using the AdamW optimizer with cosine learning rate decay.
Key architectural and optimization hyperparameters are summarized in Table~\ref{tab:hyperparameters}.

\begin{table}[!t]
\centering
\caption{Values of Key Hyperparameters\label{tab:hyperparameters}}
\begin{tabular}{@{}ll@{}}
\toprule
\textbf{Parameter} & \textbf{Value} \\ \midrule
Learning rate schedule & Cosine decay \\
Maximum learning rate & $6 \times 10^{-4}$ \\
Minimum learning rate & $6 \times 10^{-5}$ \\
Warm-up iterations & 8000 \\
Gradient accumulation steps & 16 \\
Batch size & 32 \\
\midrule
Block size & 256 \\
Encoder/Decoder layers & 3 \\
Attention heads & 3 \\
Embedding dimension ($d_{\text{embd}}$) & 192 \\
Latent dimension ($d_{\text{latent}}$) & 96 \\
Number of latent tokens ($T_{\text{latent}}$) & 32 \\
Number of communication rounds & 4 \\
\bottomrule
\end{tabular}
\end{table}

\begin{table*}[!t]
\centering
\caption{Success Rate and Number of Collisions of Different Versions of LC-MAPF on \texttt{Warehouse} Map. The provided values are average $\pm$ 95\% confidence interval. Tan boxes highlight the best mean values for visibility purposes.\label{tab:ablation}}
\resizebox{\linewidth}{!}{%
\begin{tabular}{ccccccccc}
\toprule
&  \multicolumn{8}{c}{Success rate across different LC-MAPF communication rounds} \\
\cmidrule(lr){2-9}
Agents & Rounds=1 & Rounds=2 & Rounds=3 & Rounds=4 & Rounds=5 & Rounds=6 & Rounds=7 & Rounds=8 \\
\midrule
32  & 0.000 $\pm$ 0.000 & \cellcolor{best}{1.000 $\pm$ 0.000} & \cellcolor{best}{1.000 $\pm$ 0.000} & \cellcolor{best}{1.000 $\pm$ 0.000} & \cellcolor{best}{1.000 $\pm$ 0.000} & \cellcolor{best}{1.000 $\pm$ 0.000} & \cellcolor{best}{1.000 $\pm$ 0.000} & \cellcolor{best}{1.000 $\pm$ 0.000} \\
64 & 0.000 $\pm$ 0.000 & 0.766 $\pm$ 0.070 & \cellcolor{best}{1.000 $\pm$ 0.000} & \cellcolor{best}{1.000 $\pm$ 0.000} & \cellcolor{best}{1.000 $\pm$ 0.000} & \cellcolor{best}{1.000 $\pm$ 0.000} & \cellcolor{best}{1.000 $\pm$ 0.000} & \cellcolor{best}{1.000 $\pm$ 0.000} \\
96 & 0.000 $\pm$ 0.000 & 0.094 $\pm$ 0.051 & \cellcolor{best}{1.000 $\pm$ 0.000} & \cellcolor{best}{1.000 $\pm$ 0.000} & \cellcolor{best}{1.000 $\pm$ 0.000} & \cellcolor{best}{1.000 $\pm$ 0.000} & \cellcolor{best}{1.000 $\pm$ 0.000} & \cellcolor{best}{1.000 $\pm$ 0.000} \\
128 & 0.000 $\pm$ 0.000 & 0.008 $\pm$ 0.012 & \cellcolor{best}{1.000 $\pm$ 0.000} & \cellcolor{best}{1.000 $\pm$ 0.000} & \cellcolor{best}{1.000 $\pm$ 0.000} & \cellcolor{best}{1.000 $\pm$ 0.000} & \cellcolor{best}{1.000 $\pm$ 0.000} & \cellcolor{best}{1.000 $\pm$ 0.000} \\
160 & 0.000 $\pm$ 0.000 & 0.016 $\pm$ 0.020 & 0.984 $\pm$ 0.020 & \cellcolor{best}{1.000 $\pm$ 0.000} & \cellcolor{best}{1.000 $\pm$ 0.000} & \cellcolor{best}{1.000 $\pm$ 0.000} & 0.992 $\pm$ 0.012 & \cellcolor{best}{1.000 $\pm$ 0.000} \\
192 & 0.000 $\pm$ 0.000 & 0.000 $\pm$ 0.000 & 0.742 $\pm$ 0.074 & \cellcolor{best}{0.938 $\pm$ 0.043} & 0.914 $\pm$ 0.051 & 0.883 $\pm$ 0.055 & \cellcolor{best}{0.938 $\pm$ 0.043} & \cellcolor{best}{0.938 $\pm$ 0.043} \\
\midrule
& \multicolumn{8}{c}{Collision counts across different LC-MAPF communication rounds} \\
\cmidrule(lr){2-9}
Agents & Rounds=1 & Rounds=2 & Rounds=3 & Rounds=4 & Rounds=5 & Rounds=6 & Rounds=7 & Rounds=8 \\
\midrule
32  & 299.047 $\pm$ 6.149 & 61.961 $\pm$ 2.794 & 31.977 $\pm$ 1.539 & \cellcolor{best}{5.438 $\pm$ 0.743} & 8.289 $\pm$ 1.012 & 8.055 $\pm$ 0.727 & 6.664 $\pm$ 0.754 & 7.266 $\pm$ 0.782 \\
64 & 787.773 $\pm$ 7.359 & 258.078 $\pm$ 6.611 & 127.977 $\pm$ 3.669 & \cellcolor{best}{24.914 $\pm$ 1.633} & 36.164 $\pm$ 2.379 & 42.250 $\pm$ 2.012 & 32.750 $\pm$ 2.004 & 30.188 $\pm$ 1.965 \\
96 & 1361.750 $\pm$ 14.048 & 618.430 $\pm$ 16.111 & 289.789 $\pm$ 9.089 & \cellcolor{best}{79.484 $\pm$ 4.542} & 112.781 $\pm$ 6.368 & 119.922 $\pm$ 5.572 & 105.938 $\pm$ 5.770 & 92.641 $\pm$ 5.208 \\
128 & 2077.656 $\pm$ 26.500 & 1144.828 $\pm$ 46.349 & 564.547 $\pm$ 17.383 & \cellcolor{best}{226.641 $\pm$ 13.798} & 294.492 $\pm$ 14.356 & 291.977 $\pm$ 13.338 & 264.773 $\pm$ 14.460 & 252.688 $\pm$ 12.756 \\
160 & 3317.375 $\pm$ 90.149 & 2107.891 $\pm$ 81.585 & 1057.641 $\pm$ 44.313 & \cellcolor{best}{486.711 $\pm$ 23.802} & 641.992 $\pm$ 38.691 & 639.562 $\pm$ 31.303 & 585.219 $\pm$ 28.836 & 570.234 $\pm$ 31.782 \\
192 & 5330.273 $\pm$ 142.768 & 3556.984 $\pm$ 136.774 & 2038.875 $\pm$ 107.579 & \cellcolor{best}{1175.117 $\pm$ 77.500} & 1422.742 $\pm$ 74.590 & 1408.836 $\pm$ 79.931 & 1253.203 $\pm$ 72.768 & 1224.867 $\pm$ 67.736 \\
\bottomrule
\end{tabular}
}
\end{table*}

\section{Experimental Results}

To evaluate LC-MAPF empirically, we conducted multiple series of experiments. The main series compares LC-MAPF with the state of the art in learnable MAPF, specifically MAPF-GPT~\cite{andreychuk2025mapf}, its recent fine-tuned variant MAPF-GPT-DDG~\cite{andreychuk2025advancing}, SCRIMP~\cite{wang2023scrimp}, DCC~\cite{ma2021learning}, HMAGAT~\cite{jain2026pairwise}, and MAGAT+~\cite{jain2025graph}. The original MAPF-GPT comes in three sizes: 2M, 6M, and 85M parameters. In our experiments, we used only the largest and best-performing 85M-parameter model. The experiments were conducted on the POGEMA benchmark~\cite{skrynnik2025pogema} using the same evaluation protocol as in the original MAPF-GPT paper. Specifically, we used four map types: Random, Mazes, Warehouse, and Cities Tiles. The first two are the same map types used to train LC-MAPF, whereas the latter two differ significantly in topology and are used to evaluate generalization to out-of-distribution map types. Mazes and Random maps range in size from $17\times17$ to $21\times21$ and contain up to 64 agents. To better demonstrate performance differences between the evaluated approaches, we extended the number of agents on Mazes and Random maps to 80 and 96, respectively. The Warehouse type features a single map of size $33\times46$ with restrictions on where start and goal locations can be placed (to model real-world fulfillment scenarios). The maximum number of agents on this map is 192. The Cities Tiles maps are of size $64\times64$, allowing for up to 256 agents. In all runs, the episode length was limited to 128 steps, except for Cities Tiles, where the episode length was 256. More details about the benchmark and evaluation protocol can be found in~\cite{skrynnik2025pogema}.

We also conducted additional experiments, including an ablation on the importance of communication for LC-MAPF decision making, a large-scale evaluation, an evaluation with collision shielding, and real multi-robot deployment.
By default, we do not employ collision shielding~\cite{veerapaneni2024work} for either LC-MAPF or the baselines. Although such mechanisms can efficiently post-process the actions produced by learning-based methods into collision-free joint actions, the commonly used PIBT-based shielding procedure is centralized and prioritized. This contradicts our decentralized execution setup and, more importantly, modifies the agents' original actions, making it difficult to assess the true performance of the learned policy. Therefore, the main comparison reports the performance of the learned policies without shielding. We additionally provide a separate evaluation with collision shielding enabled to analyze how this post-processing affects the relative performance of different methods.

\subsection{Comparison with the Baselines}
The first series of experimental results is shown in Fig.~\ref{fig:csr_plot} and Fig.~\ref{fig:soc_plot}. Fig.~\ref{fig:csr_plot} reports the average success rate for each map type as a function of the number of agents in the instance. LC-MAPF is on par with or better than every baseline in all evaluated cases, including the original 85M-parameter MAPF-GPT and the fine-tuned MAPF-GPT-DDG. The remaining baselines, including the recent MAGAT+ and HMAGAT methods, do not match LC-MAPF in terms of success rate.

Fig.~\ref{fig:soc_plot} presents the SoC (solution cost) ratio relative to the solution found by the centralized planner LaCAM*, using box-and-whisker plots. These results align with those in Fig.~\ref{fig:csr_plot} and show that LC-MAPF achieves the best results in most cases. Interestingly, although the original MAPF-GPT has a lower success rate than LC-MAPF on Mazes maps, its median relative solution cost is slightly better: $1.24$ versus $1.3$. However, the average relative solution cost for MAPF-GPT-85M on Mazes maps is $1.42$, compared with $1.4$ for LC-MAPF. On the Warehouse map, HMAGAT solves almost as many instances as LC-MAPF and even achieves a slightly better relative SoC ratio, but it substantially underperforms on the remaining maps.

\subsection{Communication Rounds Ablation Study}
In the LC-MAPF ablation study, we investigated how the communication mechanism affects performance. To this end, we varied the number of communication rounds used by LC-MAPF from 1 to 8. The experiments were conducted on the \texttt{Warehouse} map with the number of agents varying from 32 to 192. For each number of agents, we used all 128 testing instances provided by the POGEMA Benchmark~\cite{skrynnik2025pogema}. The episode length was set to 128. We tracked two performance indicators: success rate (the fraction of successfully solved instances) and the number of collisions. The results are shown in Table~\ref{tab:ablation}.

\begin{table*}[!t]
\centering
\caption{Communication Failure Test Results on \texttt{Random} Maps. Zero message failure means the original LC-MAPF performance. The provided values are average $\pm$ 95\% confidence interval. Best results are marked with background color per row.\label{tab:msg_failure}}
\setlength{\tabcolsep}{12pt}
\resizebox{\textwidth}{!}{%
\begin{tabular}{c cc cc cc}
\toprule
& \multicolumn{2}{c}{0\% failure}
& \multicolumn{2}{c}{20\% failure}
& \multicolumn{2}{c}{50\% failure} \\
\cmidrule(lr){2-3} \cmidrule(lr){4-5} \cmidrule(lr){6-7}
Agents & Success & Collisions & Success & Collisions & Success & Collisions \\
\midrule
8  & \cellcolor{best}{1.00 $\pm$ 0.00} & \cellcolor{best}{0.49 $\pm$ 0.22}
   & \cellcolor{best}{1.00 $\pm$ 0.00} & 0.96 $\pm$ 0.30
   & \cellcolor{best}{1.00 $\pm$ 0.00} & 1.62 $\pm$ 0.44 \\

16 & \cellcolor{best}{1.00 $\pm$ 0.00} & \cellcolor{best}{1.89 $\pm$ 0.38}
   & \cellcolor{best}{1.00 $\pm$ 0.00} & 5.49 $\pm$ 1.69
   & \cellcolor{best}{1.00 $\pm$ 0.00} & 10.27 $\pm$ 1.85 \\

24 & \cellcolor{best}{1.00 $\pm$ 0.00} & \cellcolor{best}{6.88 $\pm$ 1.65}
   & \cellcolor{best}{1.00 $\pm$ 0.00} & 12.62 $\pm$ 1.87
   & \cellcolor{best}{1.00 $\pm$ 0.00} & 29.60 $\pm$ 5.32 \\

32 & \cellcolor{best}{1.00 $\pm$ 0.00} & \cellcolor{best}{18.30 $\pm$ 4.71}
   & \cellcolor{best}{1.00 $\pm$ 0.00} & 30.98 $\pm$ 4.99
   & \cellcolor{best}{1.00 $\pm$ 0.00} & 80.79 $\pm$ 18.14 \\

48 & \cellcolor{best}{1.00 $\pm$ 0.00} & \cellcolor{best}{71.66 $\pm$ 14.07}
   & 0.98 $\pm$ 0.03 & 138.20 $\pm$ 34.15
   & 0.94 $\pm$ 0.04 & 300.72 $\pm$ 61.50 \\

64 & \cellcolor{best}{0.98 $\pm$ 0.02} & \cellcolor{best}{219.23 $\pm$ 46.80}
   & 0.93 $\pm$ 0.04 & 431.27 $\pm$ 88.40
   & 0.77 $\pm$ 0.07 & 820.64 $\pm$ 139.01 \\
\bottomrule
\end{tabular}
}
\end{table*}

\begin{table*}[!t]
\centering
\caption{Success Rates for Different Sizes of the Communication Neighborhood Evaluated on \texttt{Mazes} Maps. The Limit value shows the number of communicating agents. Limit = 13 refers to the original LC-MAPF. The provided values are average $\pm$ 95\% confidence interval. Best results are marked with background color per row.\label{tab:neighbors}}
\setlength{\tabcolsep}{12pt}
\begin{tabular}{c|ccccc}
\toprule
Agents & Limit = 1 & Limit = 2 & Limit = 4 & Limit = 8 & Limit = 13 \\
\midrule
8  & 0.96 $\pm$ 0.04 & \cellcolor{best}{1.00 $\pm$ 0.00} & \cellcolor{best}{1.00 $\pm$ 0.00} & \cellcolor{best}{1.00 $\pm$ 0.00} & \cellcolor{best}{1.00 $\pm$ 0.00}\\
16 & 0.62 $\pm$ 0.08 & 0.98 $\pm$ 0.02 & \cellcolor{best}{1.00 $\pm$ 0.00} & \cellcolor{best}{1.00 $\pm$ 0.00} & \cellcolor{best}{1.00 $\pm$ 0.00}\\
24 & 0.41 $\pm$ 0.08 & 0.91 $\pm$ 0.05 & \cellcolor{best}{1.00 $\pm$ 0.00} & \cellcolor{best}{1.00 $\pm$ 0.00} & \cellcolor{best}{1.00 $\pm$ 0.00}\\
32 & 0.28 $\pm$ 0.08 & 0.74 $\pm$ 0.07 & \cellcolor{best}{1.00 $\pm$ 0.00} & \cellcolor{best}{1.00 $\pm$ 0.00} & \cellcolor{best}{1.00 $\pm$ 0.00}\\
48 & 0.10 $\pm$ 0.05 & 0.38 $\pm$ 0.08 & 0.86 $\pm$ 0.06 & 0.96 $\pm$ 0.04 & \cellcolor{best}{0.98 $\pm$ 0.03}\\
64 & 0.06 $\pm$ 0.04 & 0.18 $\pm$ 0.07 & 0.58 $\pm$ 0.09 & 0.79 $\pm$ 0.07 & \cellcolor{best}{0.87 $\pm$ 0.06}\\
\bottomrule
\end{tabular}
\end{table*}

The obtained results clearly demonstrate that (a) LC-MAPF needs at least two communication rounds to solve at least some of the instances; (b) the best performance is achieved with the number of rounds used during training, i.e., four; and (c) further increasing the number of communication rounds does not improve the success rate, although larger numbers of rounds tend to reduce the number of collisions between agents.

\subsection{Message Failure Test}
Prior work has studied communication in Dec-POMDPs under realistic constraints such as delays, failures, and communication costs~\cite{wu2009multi,lauri2019information}. In this section, we evaluate how LC-MAPF handles message transmission failures. For each agent, with a given probability, we replace the agent's updated message at each round with a random vector sampled from the standard normal distribution. We test LC-MAPF with 20\% and 50\% message failure on a set of Random maps from the POGEMA benchmark and compare the results with the original model. The results are presented in Table~\ref{tab:msg_failure}.

The success-rate comparison reveals the negative impact of random noise messages on LC-MAPF for 48 agents and above, highlighting the importance of the information conveyed by LC-MAPF messages. However, despite the extreme setup with a 50\% message failure probability, the agents can still solve simpler tasks with up to 32 agents and achieve partial success for larger agent populations.

\subsection{Communication Bandwidth}
LC-MAPF enables each agent to receive up to 13 messages, including its own message and messages from up to 12 local neighbors. This limit was chosen because a collision is possible only with an agent that occupies one of the 12 cells closest to the current location. Communication is not strictly restricted to agents within these cells; other agents within the observable area can still participate in communication. However, agents in these cells are prioritized due to their proximity.

\begin{figure}[!t]
    \centering
    \includegraphics[width=\linewidth]{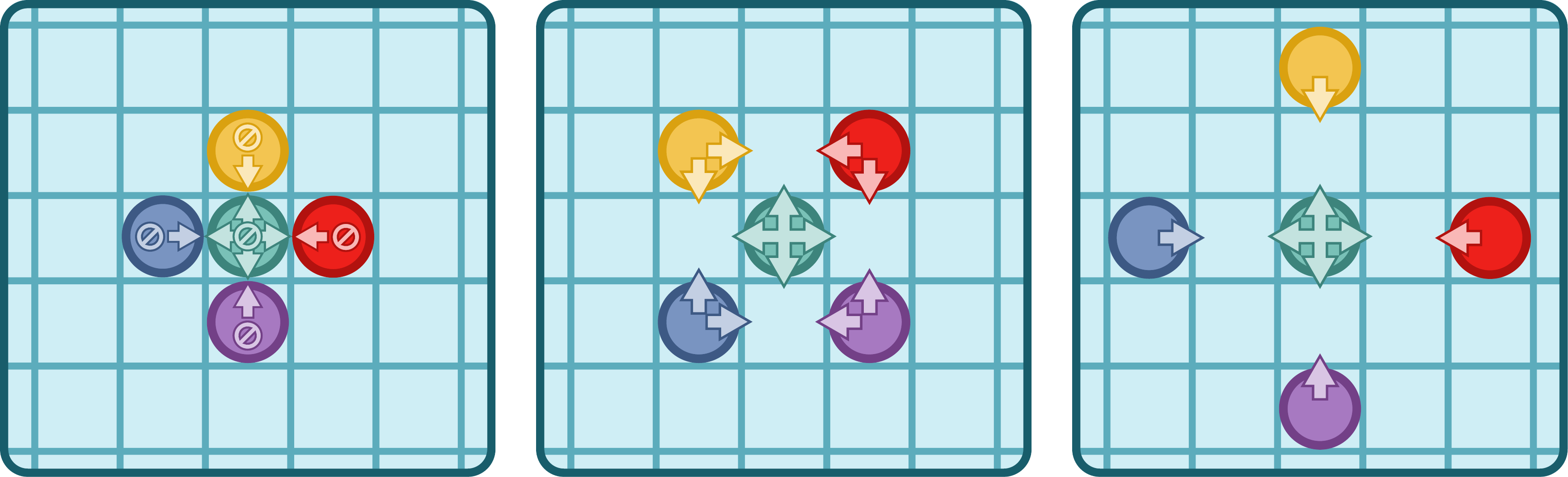}
    \caption{All agents and the corresponding actions that may result in a collision with the reference agent (marked as green).}
    \label{fig:agents_limit}
\end{figure}

For clarity, Fig.~\ref{fig:agents_limit} illustrates the relevant locations and actions that may result in a collision. The first four agents are positioned in the cardinal adjacent cells, and a collision with them is possible if both agents attempt to swap positions or if one agent chooses to wait. The next four potentially colliding agents are located in the diagonally adjacent cells; each of these has two possible actions that could lead to a collision, specifically in cases where both agents choose to enter the same cell. The final four agents are located two cells away, but a collision with them remains possible depending on the chosen actions. Regardless of the actions taken by agents in any other location, a collision with them in the current step is impossible.

We acknowledge that limited communication can affect performance. However, this constraint is motivated by practical considerations: in real-world applications, the bandwidth of communication channels is typically limited. Additionally, communication incurs costs; for instance, sending messages can significantly drain a robot's battery. Thus, there is an inherent trade-off between performance and communication bandwidth or cost.

On the one hand, in our approach, this limitation can be partially mitigated through chain communication. That is, agent $A$ may communicate with agent $B$ at communication round $t$, and agent $B$ may subsequently communicate with agent $C$ at round $t+1$. As a result, information from agent $A$ can be propagated to agent $C$ (with some delay), even though $A$ and $C$ do not directly communicate.

On the other hand, to validate the effectiveness and efficiency of the proposed neighborhood size, we test more restrictive neighborhood sizes on the Mazes maps from the POGEMA benchmark. We limit the number of communicating agents in LC-MAPF to 1, 2, 4, 8, and 13. For example, $\text{Limit}=2$ means that each agent receives messages from at most 2 agents, including itself. The agents are sorted based on their distance to the current agent. Thus, when there are 5 agents in observation, an agent receives only two messages from the closest ones. The results are presented in Table~\ref{tab:neighbors}.

The results clearly demonstrate that limiting the number of communicating agents significantly reduces the success rate, especially when strict limits (4 agents or fewer) are applied to instances with larger agent populations (48 and 64 agents). Tighter limits have little effect on instances with fewer agents because the actual number of agents present in the observations typically already satisfies the reduced limit. This experiment highlights the importance of communication for LC-MAPF and shows that restricting it can negatively affect performance.

\subsection{Large-scale Evaluation}

\begin{table}[!t]
\centering
\caption{Number of Steps Required to Solve the Corresponding Instance Depending on Obstacle Density and the Number of Agents in the Instance\label{tab:large_scale}}
\setlength{\tabcolsep}{12pt}
\begin{tabular}{c cccc}
\toprule
& \multicolumn{4}{c}{Density} \\
\cmidrule(lr){2-5}
Agents & 0\%& 10\% & 20\% & 30\% \\
\midrule
1000 & 474 & 476 & 486 & 532 \\
2000 & 456 & 497 & 495 & 2048 \\
3000 & 476 & 489 & 521 & 2048 \\
4000 & 487 & 488 & 531 & 2048 \\
5000 & 508 & 500 & 559 & 2048 \\
\bottomrule
\end{tabular}
\end{table}

The main series of experiments was conducted on the POGEMA benchmark, where instances contain at most 256 agents. Although the previous experiments already demonstrate that LC-MAPF scales linearly with the number of agents, we additionally evaluate whether LC-MAPF can scale to thousands of agents. The exponential runtime growth of other communication-based learning approaches, such as SCRIMP and DCC, prevents a direct comparison with them on large maps containing thousands of agents. We therefore evaluated LC-MAPF on 256$\times$256 random maps (with obstacle densities of 10\%, 20\%, 30\%) and empty maps (0\% density) with up to 5,000 agents. The results are shown in Table~\ref{tab:large_scale}. The reported numbers correspond to the makespan, i.e., the number of steps required for all agents to reach their goal locations and occupy them simultaneously. A value of 2048 indicates that the corresponding instance was not solved before termination. LC-MAPF maintained linear scalability in this experiment, requiring approximately 0.12 seconds per step for instances with 1,000 agents and 0.65 seconds per step for instances with 5,000 agents.

\subsection{Evaluation with Collision Shielding}

In this series of experiments, we evaluate LC-MAPF and the strongest baselines with collision-shielding post-processing enabled. This setting differs from the main experimental protocol in an important way. Without an advanced collision-shielding mechanism, collisions caused by the decentralized nature of learning-based methods are resolved by the simulator. In the POGEMA environment, this resolution is relatively simple: if an action would lead to a collision, it is replaced with a wait action. If this newly introduced wait action creates another conflict with a different agent, that agent is also forced to wait instead of executing its desired move. Thus, each collision effectively introduces a delay, since agents whose actions would cause conflicts remain in place.

In contrast, when CS-PIBT is applied, conflicts are resolved by invoking a PIBT-based procedure for the colliding agents. The method takes as input the action preferences produced by the agents and, in a centralized prioritized manner, selects a collision-free set of high-preference actions. As a result, decision making becomes a two-stage process. If the action sampled by the neural policy is collision-free, it is executed directly. Otherwise, control is transferred to CS-PIBT, which determines the final actions to be performed. This makes the reported performance depend not only on the learned policy itself, but also on the additional centralized post-processing procedure.

\begin{figure*}[t!]
    \centering
    \includegraphics[width=\textwidth]{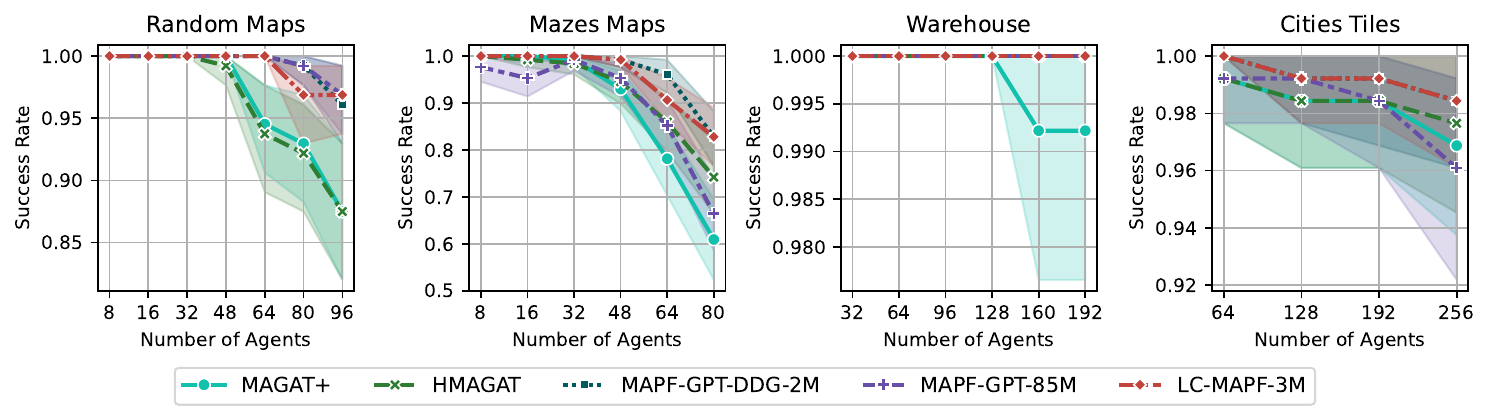}
    \caption{Success rate of LC-MAPF and the evaluated baselines with collision shielding enabled.}
    \label{fig:csr+cs}
\end{figure*}

\begin{figure*}
    \centering
    \includegraphics[width=\textwidth]{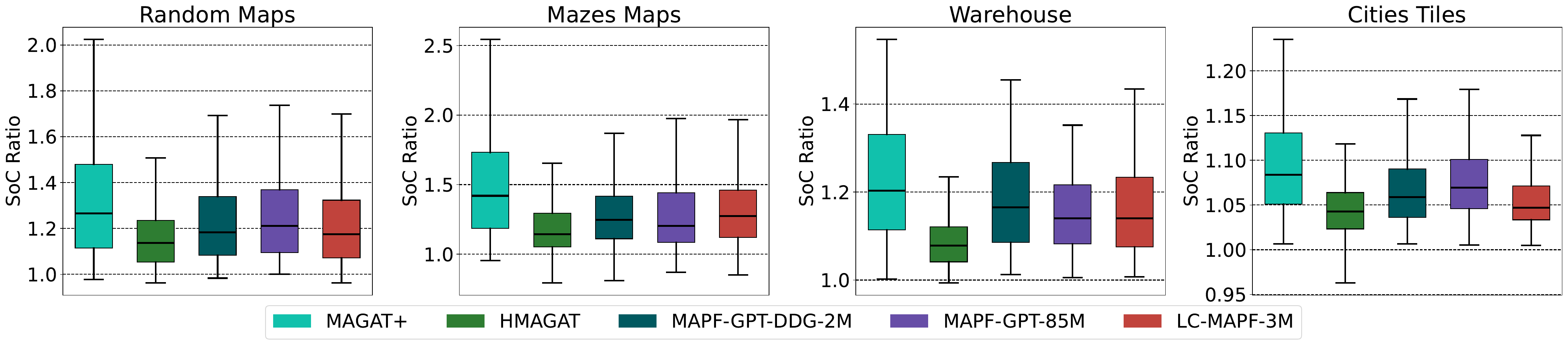}
    \caption{Relative solution cost (SoC) of LC-MAPF and the evaluated baselines with collision shielding enabled.}
    \label{fig:soc+cs}
\end{figure*}

We evaluated LC-MAPF, MAPF-GPT-85M, MAPF-GPT-DDG-2M, MAGAT+, and HMAGAT with CS-PIBT post-processing enabled. The results are shown in Fig.~\ref{fig:csr+cs} and Fig.~\ref{fig:soc+cs}. Compared with the main experiments, where the trends in success rate and solution quality were mostly consistent across map types and methods, the results with collision shielding are more difficult to interpret. For example, on Random maps, MAPF-GPT-85M achieves the highest success rate, while HMAGAT solves the fewest instances. At the same time, in terms of solution quality measured by the SoC ratio, MAPF-GPT-85M performs substantially worse than HMAGAT, which achieves the best median solution quality. Similar trends can be observed on other map types as well: HMAGAT solves fewer instances than LC-MAPF and several other baselines, but often produces solutions with better relative cost.

Overall, enabling CS-PIBT often improves the results compared with the corresponding runs without collision shielding, which is expected. However, this effect is not uniform across methods and does not always lead to better performance. For example, MAPF-GPT-85M performs worse on Mazes maps on the instances with low number of agents when CS-PIBT is enabled. A possible explanation is that MAPF-GPT uses the history of previously executed actions as part of its observation. When CS-PIBT overrides the action selected by the neural policy, the actually executed action may become inconsistent with the action distribution observed during training. This creates an out-of-distribution input for the policy at subsequent timesteps and can degrade its behavior instead of improving it.

At the same time, other methods benefit substantially from shielding. The strongest effect, especially in terms of solution quality, is observed for HMAGAT. These results show that advanced collision-resolution post-processing can significantly alter the relative performance of learning-based MAPF methods and may lead to misleading conclusions if the contribution of the learned policy is not separated from the contribution of the shielding mechanism.

\subsection{Robotics Experiments}
As an experimental validation, we reproduce a maze scenario from POGEMA on a real multi-robot platform.

\subsubsection{Maze environment}
\begin{figure}[!t]
\centering\includegraphics[height=3cm,width=0.5\textwidth,keepaspectratio,page=1]{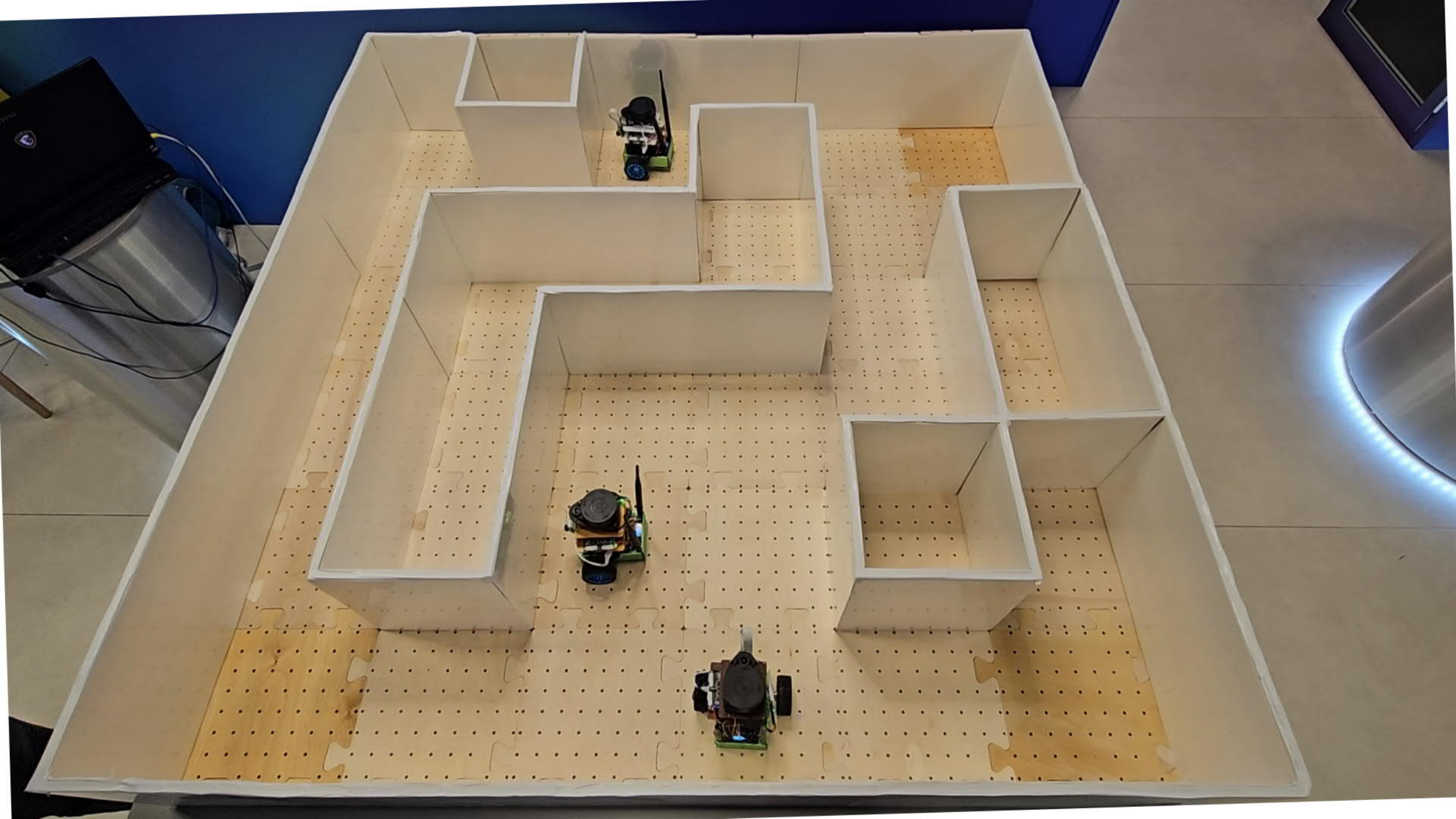}
\hspace{10px}
\centering\includegraphics[height=3cm,width=0.30\textwidth,keepaspectratio,page=2]{figures/robot_case_study.pdf}
\caption{Left: Modular and reconfigurable maze environment composed of standardized floor and wall units. Right: Deeply modified robotic agent based on Jetbot AI Kit.}
\label{fig:maze1}
\end{figure}

The physical environment is designed as a modular arena composed of standardized floor tiles and wall segments, allowing for rapid reconfiguration of the maze layout. Each floor tile is a $30~\text{cm}\times30~\text{cm}$ square plywood module with pre-drilled holes that support the attachment of wall elements. Fig.~\ref{fig:maze1} illustrates the wall placement mechanism.

\subsubsection{Robotic Agents}
Each agent is a custom-modified variant of the Waveshare JetBot platform (Fig.~\ref{fig:maze1}), significantly enhanced to support onboard perception, localization, and control. The robot dimensions are $17~\text{cm} \times 17~\text{cm} \times 22~\text{cm}$. A Jetson Nano 4GB module running Ubuntu 20.04 serves as the onboard compute unit. The robot is equipped with a pair of DC motors with encoders and an RPLIDAR A1 2D laser scanner mounted on top for mapping and obstacle detection.

Communication between the robots was implemented using the Zenoh plugin, which helps reduce the amount of data transferred over the network and the load on the Wi-Fi router. CycloneDDS was used as the ROS2 rmw implementation together with Zenoh.

For mapping, we employ SLAM Toolbox, which incrementally builds a 2D occupancy grid using onboard lidar and odometry data. Navigation is handled by the Nav2 stack, configured for both global and local planning within the constructed map. We use the ThetaStar planner and the Vector Pursuit controller.

\subsubsection{Bridging Discrete Planning and Continuous Multi-Robot Execution}
In our system, several practical challenges arise when deploying the LC-MAPF model in real robotic environments.

First, the LC-MAPF formulation operates in a discrete grid, whereas real robots function in continuous space. To bridge this gap, we estimate the coordinates of grid cell centers in the map coordinate frame. Given that the maze structure is bounded and the cell size is known, we overlay a grid of corresponding resolution onto the occupancy map. Using classical computer vision techniques, specifically contour detection, we align the grid with the map representation. This allows us to compute a transformation function between the robot map frame and the discrete model coordinate system.

Second, the LC-MAPF model assumes holonomic agents that can move between cells synchronously. In contrast, our robots are differential-drive and must first rotate toward the desired direction before executing a translation. As a result, robots would otherwise start moving at different times, potentially causing collisions. To address this, we decompose each discrete step into two sequential phases: rotation toward the target direction, during which robots wait until all agents reach the required orientation, and synchronized forward motion.

Third, the planner may assign a robot to move into a cell currently occupied by another robot. In a standard navigation stack, such a robot would be treated as a dynamic obstacle, leading to local avoidance behavior that conflicts with the global plan. However, in our setting, the planner guarantees conflict resolution over time. Therefore, we explicitly remove other robots from the local costmaps, effectively delegating collision avoidance to the planner.

Finally, there is the question of how to provide environment information to the planner. In the current implementation, we assume a static environment. Thus, the map is directly provided from the simulation, while robot poses are obtained from the TF tree under a centralized control architecture, where all robots share state information.

\subsubsection{Experimental Protocol}
The physical maze was assembled to replicate the layout used in the corresponding simulation scenario.

\begin{figure}[!t]
\centering\includegraphics[height=3.8cm,width=0.35\textwidth,keepaspectratio,page=3]{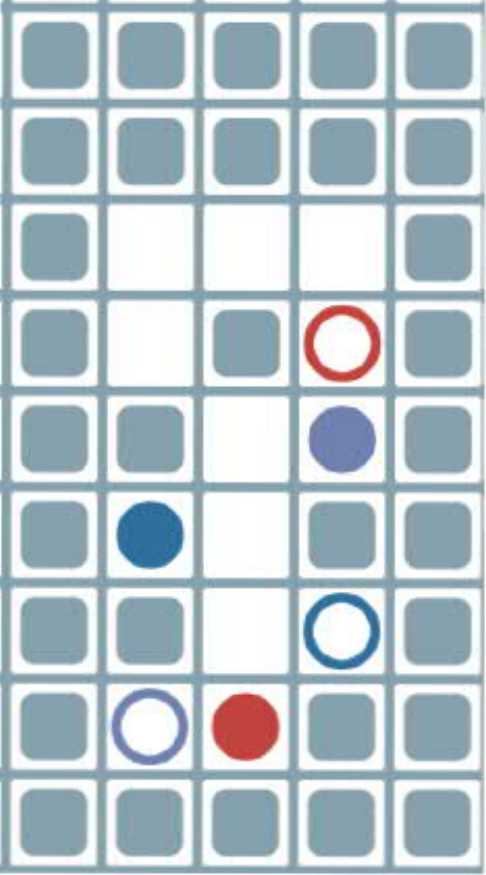}
\hspace{10px} \centering\includegraphics[height=3.8cm,width=0.53\textwidth,keepaspectratio,page=4]{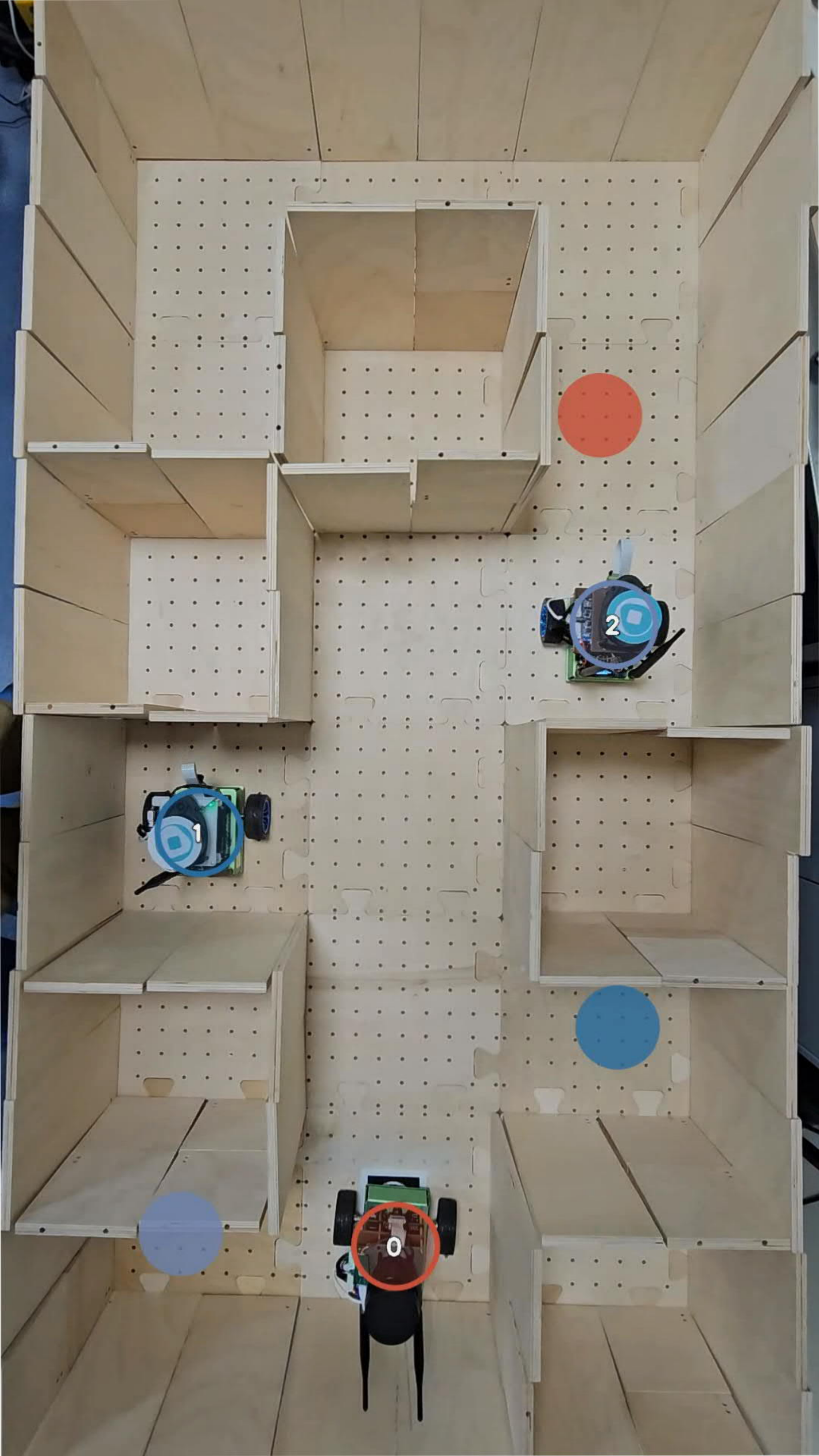}
\caption{Real-world execution of a maze scenario with 3 robots. Left: simulated environment in POGEMA. Right: corresponding physical setup used for real-world deployment.}
\label{fig:mazes_examples}
\end{figure}

An example scenario is presented in Fig.~\ref{fig:mazes_examples}, selected from several tested cases. The figure shows an instance in the POGEMA environment alongside its corresponding real-world setup. The scenario was first evaluated in simulation and then reconstructed in the physical environment using the same initial agent positions and target locations. The LC-MAPF policy was used without modification to control the agents during these real-world trials.

The resulting multi-agent behaviors, including coordination and conflict resolution, were recorded and are presented in the accompanying video materials. These experiments demonstrate that the approach is effective in the real world and that the agents can coordinate their actions to successfully reach their goals.

\section{Conclusion}

We introduced LC-MAPF, a novel communication learning framework for decentralized multi-agent pathfinding that leverages expert demonstrations without explicit communication supervision. The communication is organized in rounds to enhance cooperation between agents. Our transformer-based model outperforms state-of-the-art learning-based MAPF solvers on the POGEMA benchmark, improving coordination and cooperation across diverse scenarios.

LC-MAPF maintains linear scalability with the number of agents, overcoming a common limitation of communication-based approaches. Ablation studies confirm that multi-round local communication enhances performance without sacrificing scalability or generalization. These results highlight LC-MAPF as a generalizable pre-trained model that offers an effective and scalable solution for decentralized multi-agent pathfinding through multi-round local communication.



\bibliography{refs}

@misc{zhang2019rootmeansquarelayer,
      title={Root Mean Square Layer Normalization}, 
      author={Biao Zhang and Rico Sennrich},
      year={2019},
      eprint={1910.07467},
      archivePrefix={arXiv},
      primaryClass={cs.LG},
      url={https://arxiv.org/abs/1910.07467}, 
}

@misc{zhuo2025hybridnormstableefficienttransformer,
      title={HybridNorm: Towards Stable and Efficient Transformer Training via Hybrid Normalization}, 
      author={Zhijian Zhuo and Yutao Zeng and Ya Wang and Sijun Zhang and Jian Yang and Xiaoqing Li and Xun Zhou and Jinwen Ma},
      year={2025},
      eprint={2503.04598},
      archivePrefix={arXiv},
      primaryClass={cs.CL},
      url={https://arxiv.org/abs/2503.04598}, 
}

@misc{shazeer2020gluvariantsimprovetransformer,
      title={GLU Variants Improve Transformer}, 
      author={Noam Shazeer},
      year={2020},
      eprint={2002.05202},
      archivePrefix={arXiv},
      primaryClass={cs.LG},
      url={https://arxiv.org/abs/2002.05202}, 
}

@misc{ye2025differentialtransformer,
      title={Differential Transformer}, 
      author={Tianzhu Ye and Li Dong and Yuqing Xia and Yutao Sun and Yi Zhu and Gao Huang and Furu Wei},
      year={2025},
      eprint={2410.05258},
      archivePrefix={arXiv},
      primaryClass={cs.CL},
      url={https://arxiv.org/abs/2410.05258}, 
}

@misc{jaegle2022perceiveriogeneralarchitecture,
      title={Perceiver IO: A General Architecture for Structured Inputs \& Outputs}, 
      author={Andrew Jaegle and Sebastian Borgeaud and Jean-Baptiste Alayrac and Carl Doersch and Catalin Ionescu and David Ding and Skanda Koppula and Daniel Zoran and Andrew Brock and Evan Shelhamer and Olivier Hénaff and Matthew M. Botvinick and Andrew Zisserman and Oriol Vinyals and Joāo Carreira},
      year={2022},
      eprint={2107.14795},
      archivePrefix={arXiv},
      primaryClass={cs.LG},
      url={https://arxiv.org/abs/2107.14795}, 
}

@inproceedings{tang2024ensembling,
  title={Ensembling prioritized hybrid policies for multi-agent pathfinding},
  author={Tang, Huijie and Berto, Federico and Park, Jinkyoo},
  booktitle={2024 IEEE/RSJ International Conference on Intelligent Robots and Systems (IROS)},
  pages={8047--8054},
  year={2024},
  organization={IEEE}
}

@article{veerapaneni2024work,
  title={Work Smarter Not Harder: Simple Imitation Learning with CS-PIBT Outperforms Large Scale Imitation Learning for MAPF},
  author={Veerapaneni, Rishi and Jakobsson, Arthur and Ren, Kevin and Kim, Samuel and Li, Jiaoyang and Likhachev, Maxim},
  journal={arXiv preprint arXiv:2409.14491},
  year={2024}
}

@article{firoozi2023foundation,
  title={Foundation models in robotics: Applications, challenges, and the future},
  author={Firoozi, Roya and Tucker, Johnathan and Tian, Stephen and Majumdar, Anirudha and Sun, Jiankai and Liu, Weiyu and Zhu, Yuke and Song, Shuran and Kapoor, Ashish and Hausman, Karol and others},
  journal={The International Journal of Robotics Research},
  publisher={SAGE Publications Sage UK: London, England},
 year={2023}
}

@article{team2024octo,
  title={Octo: An open-source generalist robot policy},
  author={Team, Octo Model and Ghosh, Dibya and Walke, Homer and Pertsch, Karl and Black, Kevin and Mees, Oier and Dasari, Sudeep and Hejna, Joey and Kreiman, Tobias and Xu, Charles and others},
  journal={arXiv preprint arXiv:2405.12213},
  year={2024}
}

@inproceedings{
kim2024openvla,
title={Open{VLA}: An Open-Source Vision-Language-Action Model},
author={Moo Jin Kim and Karl Pertsch and Siddharth Karamcheti and Ted Xiao and Ashwin Balakrishna and Suraj Nair and Rafael Rafailov and Ethan P Foster and Pannag R Sanketi and Quan Vuong and Thomas Kollar and Benjamin Burchfiel and Russ Tedrake and Dorsa Sadigh and Sergey Levine and Percy Liang and Chelsea Finn},
booktitle={8th Annual Conference on Robot Learning},
year={2024},
}

@article{yang2023foundation,
  title={Foundation models for decision making: Problems, methods, and opportunities},
  author={Yang, Sherry and Nachum, Ofir and Du, Yilun and Wei, Jason and Abbeel, Pieter and Schuurmans, Dale},
  journal={arXiv preprint arXiv:2303.04129},
  year={2023}
}

@article{bommasani2021opportunities,
  title={On the opportunities and risks of foundation models},
  author={Bommasani, Rishi and Hudson, Drew A and Adeli, Ehsan and Altman, Russ and Arora, Simran and von Arx, Sydney and Bernstein, Michael S and Bohg, Jeannette and Bosselut, Antoine and Brunskill, Emma and others},
  journal={arXiv preprint arXiv:2108.07258},
  year={2021}
}

@article{fourney2024magentic,
  title={Magentic-one: A generalist multi-agent system for solving complex tasks},
  author={Fourney, Adam and Bansal, Gagan and Mozannar, Hussein and Tan, Cheng and Salinas, Eduardo and Niedtner, Friederike and Proebsting, Grace and Bassman, Griffin and Gerrits, Jack and Alber, Jacob and others},
  journal={arXiv preprint arXiv:2411.04468},
  year={2024}
}

@article{sharon2015conflict,
  title={Conflict-based search for optimal multi-agent pathfinding},
  author={Sharon, Guni and Stern, Roni and Felner, Ariel and Sturtevant, Nathan R},
  journal={Artificial intelligence},
  volume={219},
  pages={40--66},
  year={2015},
  publisher={Elsevier}
}

@inproceedings{li2022mapf,
  title={MAPF-LNS2: Fast repairing for multi-agent path finding via large neighborhood search},
  author={Li, Jiaoyang and Chen, Zhe and Harabor, Daniel and Stuckey, Peter J and Koenig, Sven},
  booktitle={Proceedings of the AAAI Conference on Artificial Intelligence},
  volume={36},
  pages={10256--10265},
  year={2022}
}

@article{ma2021learning,
  title={Learning selective communication for multi-agent path finding},
  author={Ma, Ziyuan and Luo, Yudong and Pan, Jia},
  journal={IEEE Robotics and Automation Letters},
  volume={7},
  number={2},
  pages={1455--1462},
  year={2021},
  publisher={IEEE}
}

@inproceedings{skrynnik2025pogema,
  title={{POGEMA}: A Benchmark Platform for Cooperative Multi-Agent Pathfinding},
  author={Skrynnik, Alexey and Andreychuk, Anton and Borzilov, Anatolii and Chernyavskiy, Alexander and Yakovlev, Konstantin and Panov, Aleksandr},
  year={2025},
  booktitle={The Thirteenth International Conference on Learning Representations}
}

@inproceedings{skrynnik2023learn,
  title={Learn to Follow: Decentralized Lifelong Multi-agent Pathfinding via Planning and Learning},
  author={Skrynnik, Alexey and Andreychuk, Anton and Nesterova, Maria and Yakovlev, Konstantin and Panov, Aleksandr},
  booktitle={Proceedings of the 38th AAAI Conference on Artificial Intelligence ({AAAI} 2024)},
  pages={},
  year={2024}
}

@article{sartoretti2019primal,
  title={Primal: Pathfinding via reinforcement and imitation multi-agent learning},
  author={Sartoretti, Guillaume and Kerr, Justin and Shi, Yunfei and Wagner, Glenn and Kumar, TK Satish and Koenig, Sven and Choset, Howie},
  journal={IEEE Robotics and Automation Letters},
  volume={4},
  number={3},
  pages={2378--2385},
  year={2019},
  publisher={IEEE}
}

@inproceedings{wang2023scrimp,
  title={SCRIMP: Scalable communication for reinforcement-and imitation-learning-based multi-agent pathfinding},
  author={Wang, Yutong and Xiang, Bairan and Huang, Shinan and Sartoretti, Guillaume},
  booktitle={2023 IEEE/RSJ International Conference on Intelligent Robots and Systems (IROS)},
  pages={9301--9308},
  year={2023},
  organization={IEEE}
}

@article{silver2016mastering,
  title={Mastering the game of Go with deep neural networks and tree search},
  author={Silver, David and Huang, Aja and Maddison, Chris J and Guez, Arthur and Sifre, Laurent and Van Den Driessche, George and Schrittwieser, Julian and Antonoglou, Ioannis and Panneershelvam, Veda and Lanctot, Marc and others},
  journal={nature},
  volume={529},
  number={7587},
  pages={484--489},
  year={2016},
  publisher={Nature Publishing Group}
}

@inproceedings{okumura2023lacam,
  title={Lacam: Search-based algorithm for quick multi-agent pathfinding},
  author={Okumura, Keisuke},
  booktitle={Proceedings of the AAAI Conference on Artificial Intelligence},
  volume={37},
  pages={11655--11662},
  year={2023}
}

@inproceedings{okumura2024engineering,
  title={Engineering LaCAM*: Towards Real-time, Large-scale, and Near-optimal Multi-agent Pathfinding},
  author={Okumura, Keisuke},
  booktitle={Proceedings of the 23rd International Conference on Autonomous Agents and Multiagent Systems},
  pages={1501--1509},
  year={2024}
}

@article{ruoss2024amortized,
  title={Amortized planning with large-scale transformers: A case study on chess},
  author={Ruoss, Anian and Del{\'e}tang, Gr{\'e}goire and Medapati, Sourabh and Grau-Moya, Jordi and Wenliang, Li K and Catt, Elliot and Reid, John and Lewis, Cannada A and Veness, Joel and Genewein, Tim},
  journal={Advances in Neural Information Processing Systems},
  volume={37},
  pages={65765--65790},
  year={2024}
}

@inproceedings{surynek2010optimization,
  title={An optimization variant of multi-robot path planning is intractable},
  author={Surynek, Pavel},
  booktitle={Proceedings of the 24th AAAI Conference on Artificial Intelligence ({AAAI} 2010)},
  pages={1261--1263},
  year={2010}
}

@inproceedings{stern2019multi,
  title={Multi-agent pathfinding: Definitions, variants, and benchmarks},
  author={Stern, Roni and Sturtevant, Nathan R and Felner, Ariel and Koenig, Sven and Ma, Hang and Walker, Thayne T and Li, Jiaoyang and Atzmon, Dor and Cohen, Liron and Kumar, TK Satish and others},
  booktitle={Proceedings of the 12th Annual Symposium on Combinatorial Search ({SoCS} 2019)},
  pages={151--158},
  year={2019}
}

@inproceedings{li2021lifelong,
  title={Lifelong multi-agent path finding in large-scale warehouses},
  author={Li, Jiaoyang and Tinka, Andrew and Kiesel, Scott and Durham, Joseph W and Kumar, TK Satish and Koenig, Sven},
  booktitle={Proceedings of the 35th AAAI Conference on Artificial Intelligence ({AAAI} 2021)},
  pages={11272--11281},
  year={2021}
}

@article{alkazzi2024comprehensive,
  title={A Comprehensive Review on Leveraging Machine Learning for Multi-Agent Path Finding},
  author={Alkazzi, Jean-Marc and Okumura, Keisuke},
  journal={IEEE Access},
  year={2024},
  publisher={IEEE}
}

@article{sharon2013increasing,
  title={The increasing cost tree search for optimal multi-agent pathfinding},
  author={Sharon, Guni and Stern, Roni and Goldenberg, Meir aand Felner, Ariel},
  journal={Artificial intelligence},
  volume={195},
  pages={470--495},
  year={2013},
  publisher={Elsevier}
}

@InProceedings{Wagner2011,
  author    = {Glenn Wagner and Howie Choset},
  title     = {M*: {A} complete multirobot path planning algorithm with performance bounds},
  booktitle = {Proceedings of The 2011 {IEEE/RSJ} International Conference on Intelligent Robots and Systems ({IROS} 2011)},
  pages     = {3260--3267},
  year      = {2011},
}

@inproceedings{surynek2016efficient,
  title={Efficient SAT approach to multi-agent path finding under the sum of costs objective},
  author={Surynek, Pavel and Felner, Ariel and Stern, Roni and Boyarski, Eli},
  booktitle={Proceedings of the 22nd European Conference on Artificial Intelligence ({ECAI} 2016)},
  pages={810--818},
  year={2016},
  organization={IOS Press}
}

@inproceedings{ma2019searching,
  title={Searching with consistent prioritization for multi-agent path finding},
  author={Ma, Hang and Harabor, Daniel and Stuckey, Peter J and Li, Jiaoyang and Koenig, Sven},
  booktitle={Proceedings of the AAAI conference on artificial intelligence},
  volume={33},
  pages={7643--7650},
  year={2019}
}

@article{skrynnik2023switch,
  title={When to switch: planning and learning for partially observable multi-agent pathfinding},
  author={Skrynnik, Alexey and Andreychuk, Anton and Yakovlev, Konstantin and Panov, Aleksandr I},
  journal={IEEE Transactions on Neural Networks and Learning Systems},
  year={2023},
  publisher={IEEE}
}

@inproceedings{li2023intersection,
  title={Intersection coordination with priority-based search for autonomous vehicles},
  author={Li, Jiaoyang and Lin, Eugene and Vu, Hai L and Koenig, Sven and others},
  booktitle={Proceedings of the 37th AAAI Conference on Artificial Intelligence ({AAAI} 2023)},
  pages={11578--11585},
  year={2023}
}

@inproceedings{andreychuk2025mapf,
  title={{MAPF-GPT}: Imitation learning for multi-agent pathfinding at scale},
  author={Andreychuk, Anton and Yakovlev, Konstantin and Panov, Aleksandr and Skrynnik, Alexey},
  booktitle={Proceedings of the AAAI Conference on Artificial Intelligence},
  volume={39},
  pages={23126--23134},
  year={2025}
}

@InProceedings{standley2010finding,
  author    = {T. S. Standley},
  title     = {Finding optimal solutions to cooperative pathfinding problems},
  booktitle = {Proceedings of The 24th AAAI Conference on Artificial Intelligence ({AAAI} 2010)},
  year      = {2010},
  pages     = {173-178},
}

@article{okumura2022priority,
  title={Priority inheritance with backtracking for iterative multi-agent path finding},
  author={Okumura, Keisuke and Machida, Manao and D{\'e}fago, Xavier and Tamura, Yasumasa},
  journal={Artificial Intelligence},
  volume={310},
  pages={103752},
  year={2022},
  publisher={Elsevier}
}

@article{li2021message,
  title={Message-aware graph attention networks for large-scale multi-robot path planning},
  author={Li, Qingbiao and Lin, Weizhe and Liu, Zhe and Prorok, Amanda},
  journal={IEEE Robotics and Automation Letters},
  volume={6},
  number={3},
  pages={5533--5540},
  year={2021},
  publisher={IEEE}
}

@inproceedings{liu2020mapper,
  title={Mapper: Multi-agent path planning with evolutionary reinforcement learning in mixed dynamic environments},
  author={Liu, Zuxin and Chen, Baiming and Zhou, Hongyi and Koushik, Guru and Hebert, Martial and Zhao, Ding},
  booktitle={Proceedings of the 2020 {IEEE/RSJ} International Conference on Intelligent Robots and Systems ({IROS} 2020)},
  pages={11748--11754},
  year={2020},
  organization={IEEE}
}

@inproceedings{ma2021distributed,
  title={Distributed heuristic multi-agent path finding with communication},
  author={Ma, Ziyuan and Luo, Yudong and Ma, Hang},
  booktitle={2021 IEEE International Conference on Robotics and Automation ({ICRA} 2021)},
  pages={8699--8705},
  year={2021},
  organization={IEEE}
}

@inproceedings{honig2016multi,
  title={Multi-Agent Path Finding with Kinematic Constraints.},
  author={H{\"o}nig, Wolfgang and Kumar, TK Satish and Cohen, Liron and Ma, Hang and Xu, Hong and Ayanian, Nora and Koenig, Sven},
  booktitle={Proceedings of The 26th International Conference on Automated Planning and Scheduling ({ICAPS} 2016)},
  pages={477--485},
  year={2016}
}

@inproceedings{ma2019lifelong,
	author = {H. Ma and W. H\"{o}nig and T. K. S. Kumar and N. Ayanian and S. Koenig},
	title = {Lifelong Path Planning with Kinematic Constraints for Multi-Agent Pickup and Delivery},
	booktitle = {Proceedings of the 33rd AAAI Conference on Artificial Intelligence ({AAAI} 2019)},
  pages={7651--7658},
	year = 2019
}

@inproceedings{veerapaneni2024improving,
  title={Improving learnt local MAPF policies with heuristic search},
  author={Veerapaneni, Rishi and Wang, Qian and Ren, Kevin and Jakobsson, Arthur and Li, Jiaoyang and Likhachev, Maxim},
  booktitle={Proceedings of the International Conference on Automated Planning and Scheduling},
  volume={34},
  pages={597--606},
  year={2024}
}

@inproceedings{wang2025lns2,
  title={LNS2+ RL: Combining multi-agent reinforcement learning with large neighborhood search in multi-agent path finding},
  author={Wang, Yutong and Duhan, Tanishq and Li, Jiaoyang and Sartoretti, Guillaume},
  booktitle={Proceedings of the AAAI Conference on Artificial Intelligence},
  volume={39},
  pages={23343--23350},
  year={2025}
}

@article{berner2019dota,
  title={Dota 2 with large scale deep reinforcement learning},
  author={Berner, Christopher and Brockman, Greg and Chan, Brooke and Cheung, Vicki and D{k{e}}biak, Przemys{l}aw and Dennison, Christy and Farhi, David and Fischer, Quirin and Hashme, Shariq and Hesse, Chris and others},
  journal={arXiv preprint arXiv:1912.06680},
  year={2019}
}

@misc{sagirova2025srmt,
      title={SRMT: Shared Memory for Multi-agent Lifelong Pathfinding}, 
      author={Alsu Sagirova and Yuri Kuratov and Mikhail Burtsev},
      year={2025},
      eprint={2501.13200},
      archivePrefix={arXiv},
      primaryClass={cs.LG},
      url={https://arxiv.org/abs/2501.13200}, 
}

@article{phan2025generative,
  title={Generative Curricula for Multi-Agent Path Finding via Unsupervised and Reinforcement Learning},
  author={Phan, Thomy and Phan, Timy and Koenig, Sven},
  journal={Journal of Artificial Intelligence Research},
  volume={82},
  pages={2471--2534},
  year={2025}
}

@inproceedings{wu2009multi,
  title={Multi-agent online planning with communication},
  author={Wu, Feng and Zilberstein, Shlomo and Chen, Xiaoping},
  booktitle={Proceedings of the International Conference on Automated Planning and Scheduling},
  volume={19},
  pages={321--328},
  year={2009}
}

@inproceedings{lauri2019information,
  author = {Lauri, Mikko and Pajarinen, Joni and Peters, Jan},
  title = {Information Gathering in Decentralized POMDPs by Policy Graph Improvement},
  year = {2019},
  isbn = {9781450363099},
  publisher = {International Foundation for Autonomous Agents and Multiagent Systems},
  address = {Richland, SC},
  booktitle = {Proceedings of the 18th International Conference on Autonomous Agents and Multiagent Systems (AAMAS)},
  pages = {1143–1151},
  numpages = {9},
  location = {Montreal QC, Canada},
  series = {AAMAS ’19},
  codelink = {https://github.com/laurimi/npgi},
  archiveprefix = {arXiv},
  eprint = {1902.09840},
  url = {https://dl.acm.org/doi/abs/10.5555/3306127.3331815}
}

@inproceedings{andreychuk2025advancing,
  title={Advancing Learnable Multi-Agent Pathfinding Solvers with Active Fine-Tuning},
  author={Andreychuk, Anton and Yakovlev, Konstantin and Panov, Aleksandr and Skrynnik, Alexey},
  booktitle={2025 IEEE/RSJ International Conference on Intelligent Robots and Systems (IROS)},
  pages={10564--10571},
  year={2025},
  organization={IEEE}
}

@article{jain2026pairwise,
  title={Pairwise is Not Enough: Hypergraph Neural Networks for Multi-Agent Pathfinding},
  author={Jain, Rishabh and Okumura, Keisuke and Amir, Michael and Lio, Pietro and Prorok, Amanda},
  journal={arXiv preprint arXiv:2602.06733},
  year={2026}
}

@article{jain2025graph,
  title={Graph Attention-Guided Search for Dense Multi-Agent Pathfinding},
  author={Jain, Rishabh and Okumura, Keisuke and Amir, Michael and Prorok, Amanda},
  journal={arXiv preprint arXiv:2510.17382},
  year={2025}
}

\end{document}